\documentclass{article}

     \usepackage[preprint]{neurips_2022}

\usepackage[utf8]{inputenc} 
\usepackage[T1]{fontenc}    
\usepackage{url}            
\usepackage{booktabs}       
\usepackage{amsfonts}       
\usepackage{nicefrac}       
\usepackage{microtype}      
\usepackage{xcolor}         

\usepackage{graphicx}
\usepackage{subcaption}
\usepackage{todonotes}
\usepackage{epigraph}

\usepackage{hyperref}
\hypersetup{
    colorlinks=true,
    linkcolor=red,
    filecolor=magenta,      
    urlcolor=blue,
    citecolor=purple,
    pdftitle={Overleaf Example},
    pdfpagemode=FullScreen,
    }

\usepackage{xparse}
\NewDocumentCommand{\codeword}{m}{%
    \texttt{\textcolor{blue}{#1}}%
}

\begin{document}

\title{Towards Compute-Optimal Transfer Learning}

%

\author{%
  Massimo Caccia \thanks{Current affiliation: Mila - Quebec AI Institute. Work done while interning at DeepMind. email: \texttt{massimo.p.caccia@gmail.com}} \\
   \And
   Alexandre Galashov\thanks{DeepMind, London} \\
   \AND
   Arthur Douillard\footnotemark[2] \\
   \And
   Amal Rannen-Triki\footnotemark[2]\\
   \And
   Dushyant Rao\footnotemark[2] \\
  \And
  Michela Paganini\footnotemark[2]\\
  \And
  Laurent Charlin\thanks{Mila, HEC Montréal, Canada CIFAR AI Chair}\\
  \And
  Marc'aurelio Ranzato\footnotemark[2]\\
  \And
  Razvan Pascanu\footnotemark[2]\\
}

\maketitle

\begin{abstract}

    The field of transfer learning is undergoing a significant shift with the introduction of large pretrained models which have demonstrated strong adaptability to a variety of downstream tasks. However, the high computational and memory requirements to finetune or use these models can be a hindrance to their widespread use.
    In this study, we present a solution to this issue by proposing a simple yet effective way to trade computational efficiency for \emph{asymptotic} performance which we define as the performance a learning algorithm achieves as compute tends to infinity.  Specifically, we argue that zero-shot structured pruning of pretrained models allows them
    to increase compute efficiency with minimal reduction in performance.
    We evaluate our method on the Nevis'22 continual learning benchmark that offers a diverse set of transfer scenarios.
    Our results show that pruning convolutional filters of pretrained models can lead to more than 20\% performance improvement in low computational regimes.
\end{abstract}



\section{Asymptotic Performance versus Computational Efficiency}

\begin{figure}
    \centering
    \begin{subfigure}[b]{0.32\textwidth}
        \includegraphics[trim={3cm 0 0 0}, width=\textwidth]{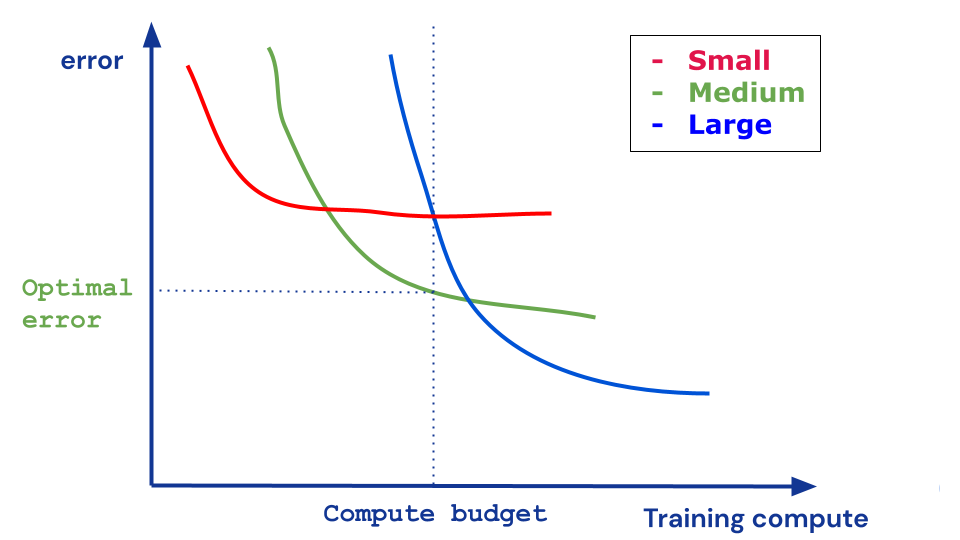}
        \caption{}
        \label{fig:fig1a}
    \end{subfigure}
    \begin{subfigure}[b]{0.32\textwidth}
        \includegraphics[trim={3cm 0 0 0}, width=\textwidth]{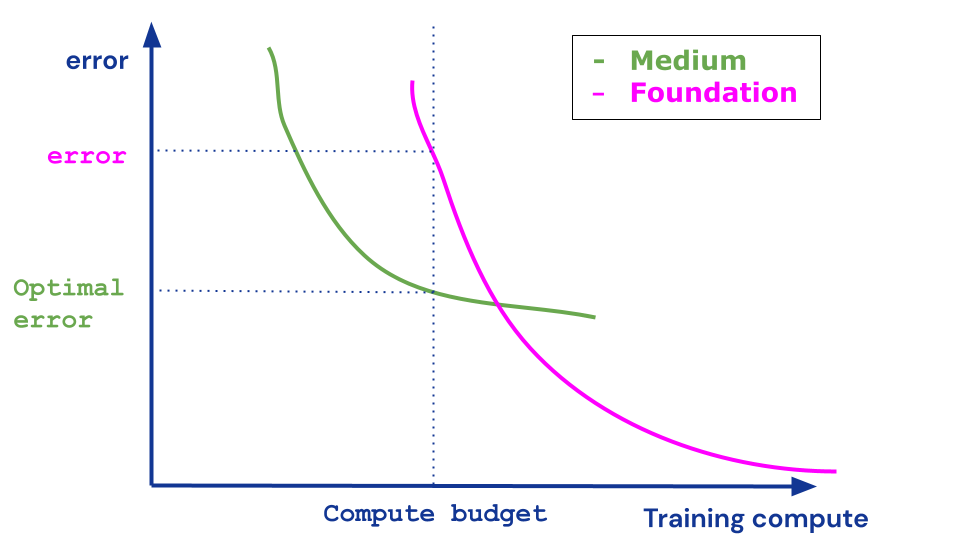}
        \caption{}
        \label{fig:fig1b}
    \end{subfigure}
    \begin{subfigure}[b]{0.32\textwidth}
        \includegraphics[trim={3cm 0 0 0}, width=\textwidth]{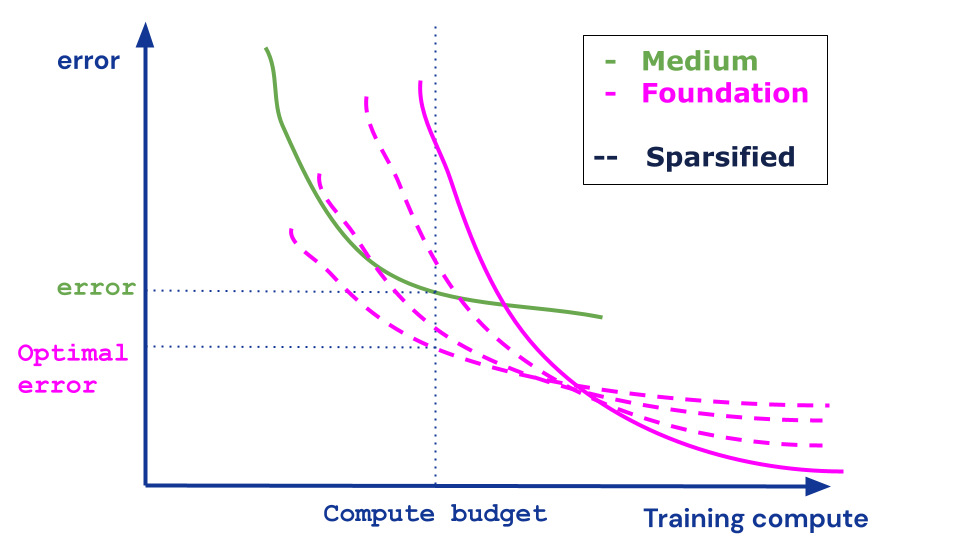}
        \caption{}
        \label{fig:fig1c}
    \end{subfigure}
    \caption{ Discover the ideal model architecture and size that best fits your
        target data and training compute budget. The efficiency of different
        model architectures/sizes varies in terms of asymptotic performance vs compute
        efficiency. The \textbf{left plot} shows the goal of finding a model
        that optimally fits your compute budget, but a pretrained foundation
        model is a recommended approach. The \textbf{middle plot} highlights
        that the computational complexity of the foundation model may exceed
        your budget. The \textbf{right plot} presents our solution: zero-shot
        structured pruning, which increases computational efficiency at the cost
        of reduced asymptotic performance, allowing the foundation model to fit your
        compute budget. }
    \label{fig:fig1}
\end{figure}



An additional important aspect is that unstructured sparsity while in theory should save compute, it rarely can be exploited
currently, on traditional hardware and due to the lack of low-level efficient sparse operations. 
To alleviate this, we focus on structured sparsity that can be readily lead to reduction in computation. 
We show that depending on the computational budget, there exists a sparsity level that 
will lead to a better performance/compute tradeoff then finetuning the original dense model.

Structured pruning is a popular technique to reduce the size of deep neural
networks and improve computational efficiency. One of the most commonly used
structured pruning approaches is channel pruning \citep{he2017channel}, which reduces the number of
channels in a layer. The computational gains from channel pruning are
substantial, as the FLOPs decrease at the square of the pruning rate. However,
as the representations become smaller, the model has less capacity for learning.

An alternative approach to channel pruning is convolution filter pruning \citep{li2016pruning}, which
prunes at a finer granularity. With convolution filter pruning, the
representation size is preserved, but more aggressive pruning is required to
achieve the same computational speedup. This is because there are fewer
activations to connect between layers in channel pruning. Therefore, for a fixed
sparsity value, convolutional pruning enjoys more representations but channel
more weights. 

Experiments will help us understand the relative merits of channel pruning and
convolution filter pruning. For a visual comparison, see Appendix
\ref{app:pruning_visual}.

\section{Related Work} 

The closest work to ours is \cite{chen2021lottery}, which examines the existence
of sparse subnetworks in \emph{dense} pretrained computer vision models using
the lottery ticket hypothesis. With unstructured iterative magnitude pruning
(IMP) they find sparse subnetworks that retain the transfer performance of full
models. Our work extends this by using structured pruning for computational
speedup, zero-shot pruning to save compute, and empirically examining the
tradeoff between performance and compute.

In the field of transferring sparse pretrained models, several works have
explored the effect of different unstructured pruning techniques during
pretraining on transfer performance
\citep{mehta2019sparse,morcos2019one,paganini2020bespoke,sabatelli2020transferability,Sun2022DiSparseDS,liu-etal-2022-learning-win}.
Another line of work focuses on pruning channels \citep{he2017channel} and
convolution filters \citep{li2016pruning}. Additionally, structured IMP has been
studied in \cite{chen2022coarsening,paganini2020iterative,rachwan2022winning}.

Another work that is relevant to our paper is \cite{frantar2023massive}, which
structurally sparsifies a pretrained GPT without much loss in performance, We
refer to \cite{hoefler2021sparsity} for a thourought investigation of sparse
neural networks.

\section{Experiments}

\subsection{Experimental Details}

In order to study methods in both compute and performance space, we conduct
experiments using the new Nevis'22 benchmark \citep{bornschein2022nevis}. It consists of
a stream of over 100 visual classification tasks, sorted chronologically and
extracted from papers sampled from computer vision proceedings spanning the last
three decades. The stream reflects the research community's understanding of
meaningful tasks at any given point in time, and while limited to
classification, it includes a diverse range of tasks including OCR, texture
analysis, crowd counting, scene recognition, and more. The diversity of tasks is
also reflected in the range of dataset sizes, which span over four orders of
magnitude.

For our experiments, we focus on the SHORT stream version of Nevis, which
consists of 25 tasks, including 16 object recognition tasks and 8 non-object
tasks (OCR, scenes, faces, texture, medical). To evaluate approaches that do not
start from scratch, Nevis'22 prescribes ImageNet pretraining. Accordingly, we
have removed ImageNet from the SHORT stream. The non-object tasks are further
considered out-of-distribution (OoD) with respect to ImageNet pretraining.

The backbone of our model is a ResNet50 architecture, taken from the model zoo, a platform of pretrained deep learning models. The ResNet is composed of four ResNet blocks which we label 0 to 3. For the
majority of our experiments, we perform pruning locally, i.e., we prune the
smaller weights of individual layers. We prune such that the resulting model achieves a desired fraction of the initial required FLOPs required for the forward and backward passes of the model. For weight and convolutional pruning, a sparsity of X\% means that X\% of the weights or convolution filters have been removed, resulting in only 1-X\% of the initial FLOPs. For channel pruning, we instead remove $1-\sqrt{1-X\%}$ because the FLOPs decrease at the square of the pruning rate


\subsection{Zero-shot structured pruning of dense pretrained models efficiently trades-off performance
for compute}

\begin{figure*}[t]
    \centering
    \begin{subfigure}[b]{0.475\textwidth}
        \centering
        \includegraphics[width=\textwidth]{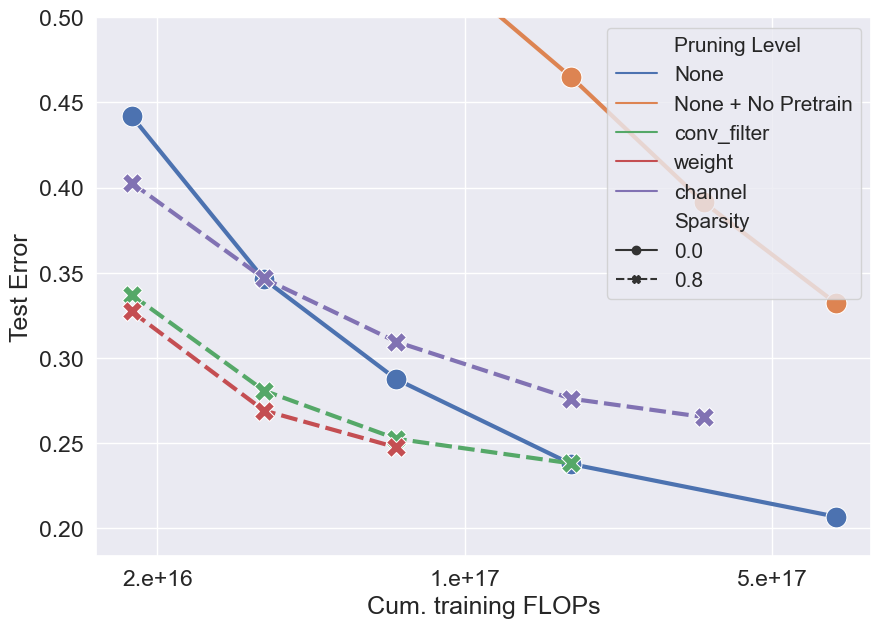}
        \caption[]%
        {{\small Transfer learning -- Object}}    
        \label{fig:lvl_ablation_independent_object}
    \end{subfigure}
    \hfill
    \begin{subfigure}[b]{0.475\textwidth}  
        \centering 
        \includegraphics[width=\textwidth]{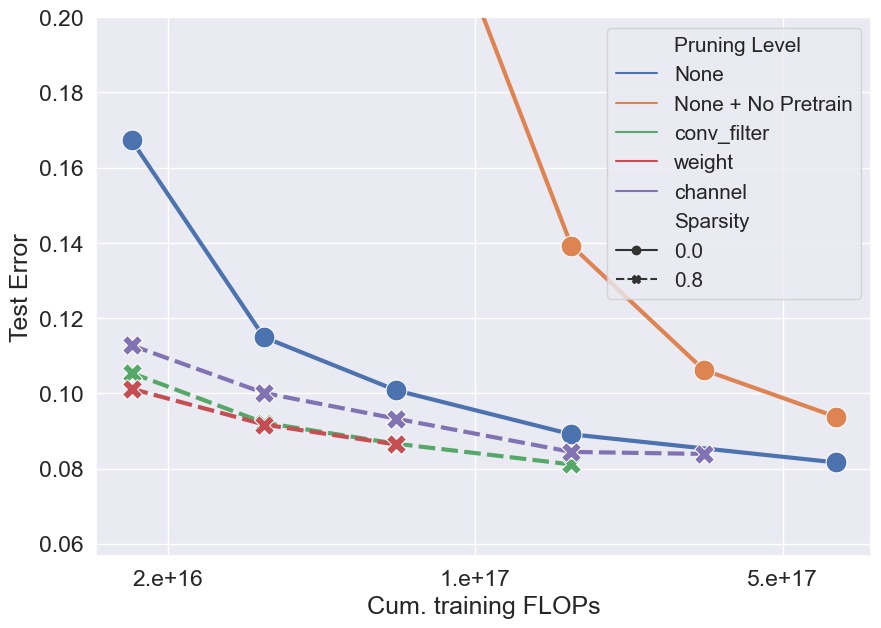}
        \caption[]%
        {{\small Transfer learning -- Non-object}}    
        \label{fig:lvl_ablation_independent_non_object}
    \end{subfigure}
    \vskip\baselineskip
    \begin{subfigure}[b]{0.475\textwidth}   
        \centering 
        \includegraphics[width=\textwidth]{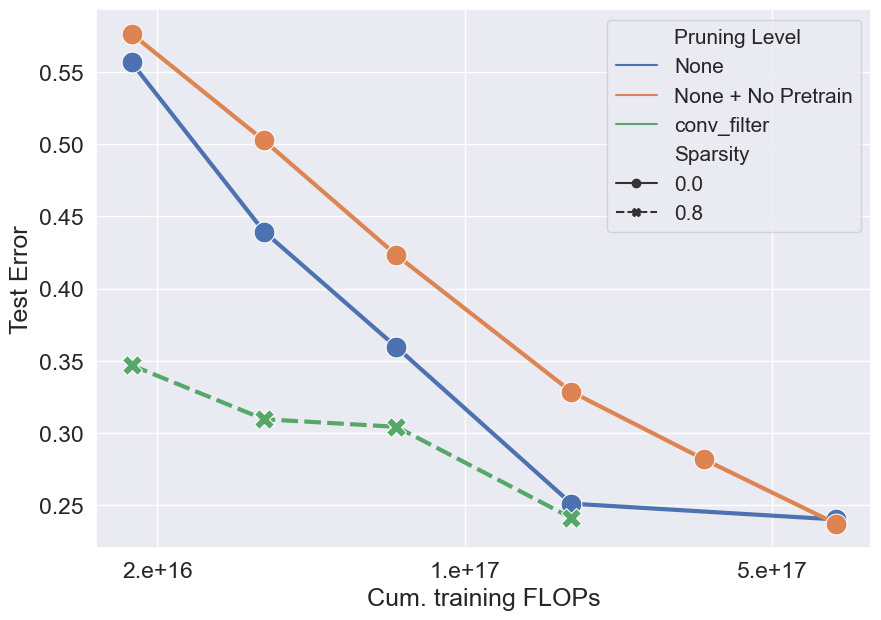}
        \caption[]%
        {{\small Continual learning -- Object}}    
        \label{fig:lvl_ablation_previous_object}
    \end{subfigure}
    \hfill
    \begin{subfigure}[b]{0.475\textwidth}   
        \centering 
        \includegraphics[width=\textwidth]{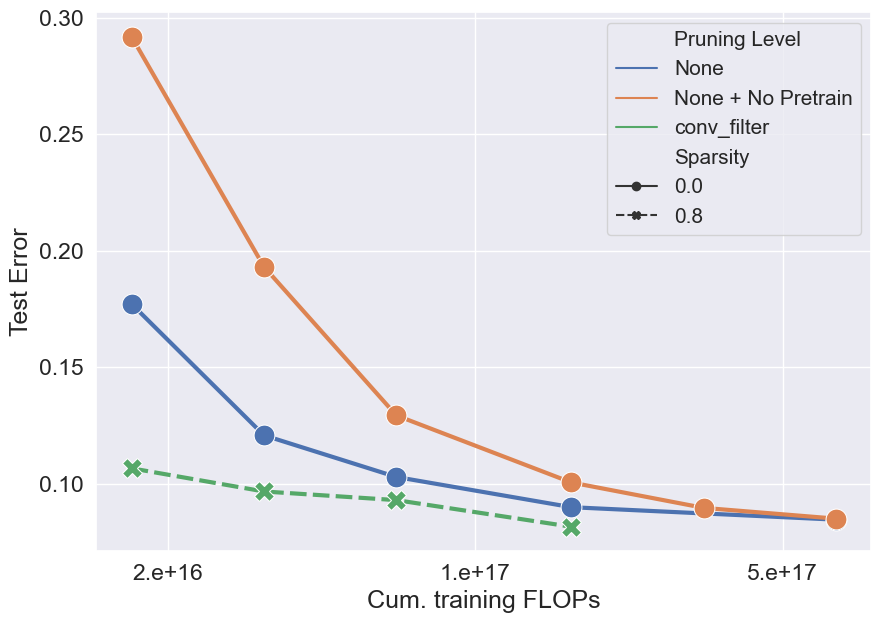}
        \caption[]%
        {{\small Continual learning -- Non-bject}}    
        \label{fig:lvl_ablation_previous_non_object}
    \end{subfigure}
    \caption{\small 
        \textbf{Zero-shot structure pruning of dense pretrained models efficiently trades-off 
                performance for compute.} In the first set of experiments, we compare 
                different pruning levels on the same pretrained model in transfer learning (top 
                plots). Unstructured Weight pruning achieves the desired tradeoff, but doesn't offer 
                any computational gains on accelerators. Channel pruning is not so efficient in the 
                tradeoff. Alluringly, convolutional pruning achieves almost the same tradeoff as 
                weight pruning. In the second set of experiments (bottom plots), we compare 
                compare convolutional pruning with finetuning the full model in the continual 
                learning setting. Convolutional pruning is again an effective solution. 
    } 
    \label{fig:level_ablation}
\end{figure*}

In our experiments, we first evaluate the performance of structured pruning in
the transfer learning scenario. The results, as shown in Figures
\ref{fig:lvl_ablation_independent_object} and
\ref{fig:lvl_ablation_independent_non_object}, indicate that convolutional filter
pruning performs similarly to unstructured weight pruning, where the latter does
not lead to computational efficiency gains on accelerators. Furthermore,
compared to channel pruning, convolutional filter pruning is the most efficient
way to prune the dense pretrained model. This is especially true for transfer
learning tasks that are closer to the pretraining distribution, i.e. objects. It
is possible that for target distributions that are closer to the pretraining
distribution, more of the learned features are transferable, and therefore, it
is necessary to keep all of them.

Based on these results, we choose to stick with convolutional pruning for the
remaining experiments. We also evaluate the performance of convolutional pruning
in the continual learning scenario (see Figure
\ref{fig:lvl_ablation_previous_object} and
\ref{fig:lvl_ablation_independent_non_object}) and find that it provides an
efficient way to trade off performance for compute.

\subsection{Zero-shot structured pruning is better than other approaches}

There are different approaches to reducing the computational complexity of a
pretrained model in transfer learning. The first set of baselines involves
finetuning only the last layers of the pretrained model. This approach saves
computation by only performing backpropagation on the last layers, which is
twice as expensive as the forward pass. The notation \codeword{ft\_block\_Xto3}
means finetuning the ResNet blocks X to 3 (the last one), and visual support can
be found in the Appendix \ref{app:paramefficient_visual}.

\begin{figure*}[t]
    \centering
    \begin{subfigure}[b]{0.475\textwidth}
        \centering
        \includegraphics[width=\textwidth]{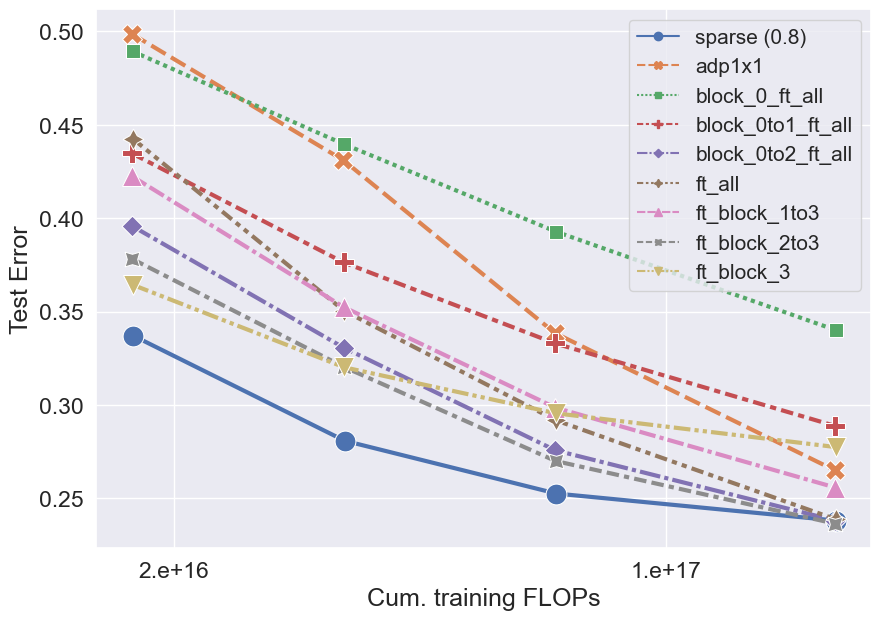}
        \caption[]%
        {{\small Transfer learning -- Object}}    
        \label{fig:mean and std of net14}
    \end{subfigure}
    \hfill
    \begin{subfigure}[b]{0.475\textwidth}  
        \centering 
        \includegraphics[width=\textwidth]{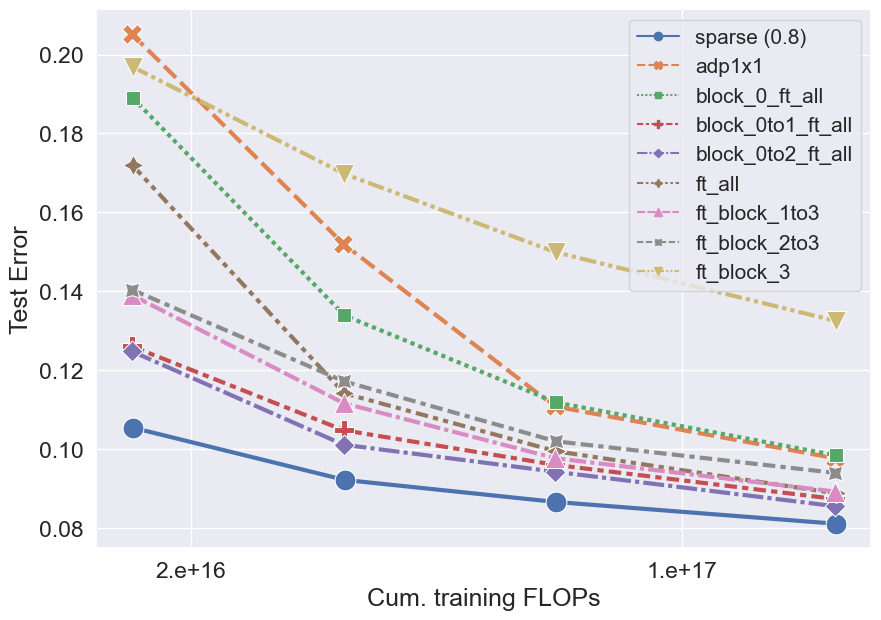}
        \caption[]%
        {{\small Transfer learning -- Non-object}}    
        \label{fig:mean and std of net24}
    \end{subfigure}
    \vskip\baselineskip
    \begin{subfigure}[b]{0.475\textwidth}   
        \centering 
        \includegraphics[width=\textwidth]{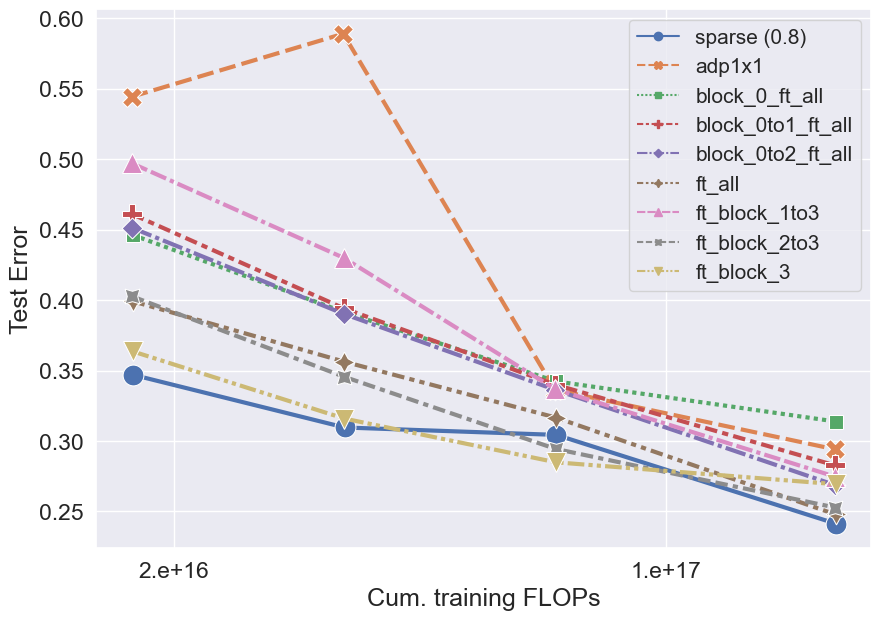}
        \caption[]%
        {{\small Continual learning -- Object}}    
        \label{fig:mean and std of net34}
    \end{subfigure}
    \hfill
    \begin{subfigure}[b]{0.475\textwidth}   
        \centering 
        \includegraphics[width=\textwidth]{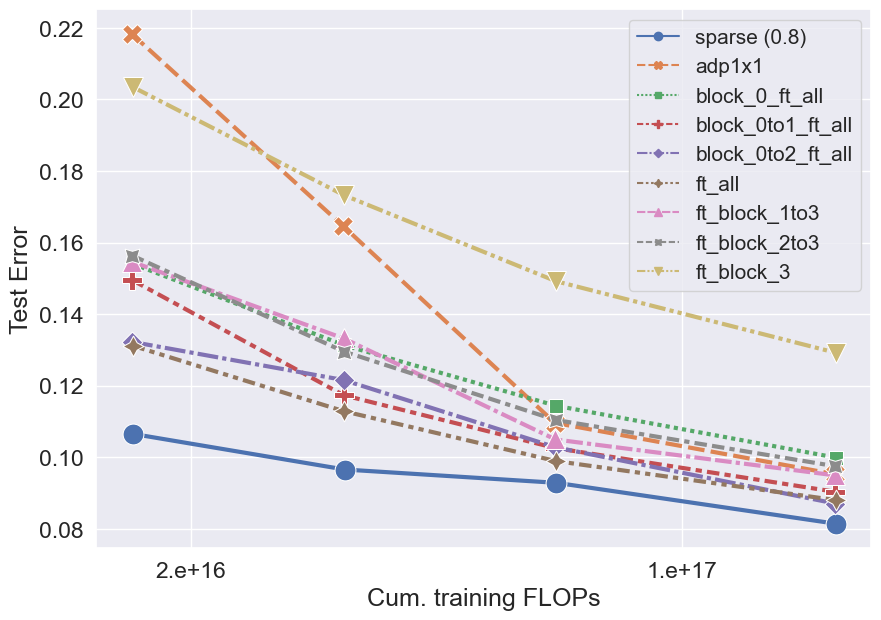}
        \caption[]%
        {{\small Continual learning -- Non-bject}}    
        \label{fig:mean and std of net44}
    \end{subfigure}
    \caption{\small
    Our results demonstrate that \textbf{the zero-shot structured pruning approach is more effective in balancing the trade-off between performance and compute compared to other methods.} In both object and non-object recognition tasks and for transfer and continual learning scenarios, the zero-shot structured pruning approach exhibits superior efficiency. The pruning approach benefits from both full modification of the feature representations and the expressivity of the pretrained model, whereas the baselines are limited to one or the other.
    } 
    \label{fig:paramefficient}
\end{figure*}

The results are presented in Figure \ref{fig:paramefficient}. The proposed
method outperforms the compared baselines in transfer learning tasks across
different datasets, especially in the object recognition domain. This is because
the baselines cannot modify the initial feature representation, which may
require more adaptation as the target data distribution deviates from the
pre-training distribution.

However, the pruning approach does not achieve superior performance over the
baselines in the continual learning scenario on object recognition datasets.
This outcome is expected, as the ImageNet pre-training already provides the
model with informative representations, leaving limited scope for knowledge
transfer between tasks. In this case, it may be more effective to freeze the
early layers to avoid catastrophic forgetting.

The second set of baselines involves selecting only the first layers and training
everything. This approach can modify the earlier representations of the data
but does not have the same depth as other approaches, and thus incurs a loss of
expressivity. The notation \codeword{block\_0toX\_ft\_all} means picking the first
X blocks and finetuning everything, with visual support available in the
Appendix \ref{app:paramefficient_visual}.

The results are again displayed in Figure \ref{fig:paramefficient}. The
performance of the \codeword{block\_0toX\_ft\_all} baselines is inferior to that of
the \codeword{ft\_block\_Xto3}. The proposed zero-shot structured pruning approach
demonstrates favorable results across all scenarios. Although the baselines can
modify the initial feature representations, the loss in expressivity may render
them less effective.

The final baseline utilizes  1x1 convolutional adapters
\citep{rebuffi2017learning}. At first glance, the use of adapters may appear as
a suboptimal trade-off between performance and compute, as they do
not reduce the computational demands of the forward and backward passes.
However, some literature suggests that this approach may not be computationally
efficient \cite{}. Our experiments confirm that adapters (denoted as
\codeword{adp1x1}) are not an effective solution for the studied use case, which
is unsurprising.

\paragraph{Ablations} We have conducted further empirical analysis, namely that of the relationship between pruning rate and the trade-off between performance and compute (refer to Appendix \ref{app:sparsity_vals}). Additionally, we found that the choice of pruning signal is crucial to the performance of the method, with pruning based on the magnitude of the weights being more effective than random pruning (see Appendix \ref{app:sparsity_stage}).


\section{Future Work \& Discussion}

The study of scaling laws, which typically explores the relationship between
computation, data, and model size, is a topic of ongoing research. The
development of scaling laws for sparsity and transfer would have the potential
to predict the ideal level of sparsity for a specific transfer task and
computational budget. Another avenue of research is adaptive pruning, where the pruning process is dynamically adjusted during training via the target loss. Further investigation is also needed to better understand the effects of pretraining on sparsity, including the number
of pretraining steps, supervised versus self-supervised training, and model
size.

Comparing the results of structured pruning with other methods such as
distillation is crucial in evaluating its effectiveness and limitations. Another
promising area for future investigation is the use of zero-shot structured
pruning in the context of pretrained transformers for natural language
processing. In this case, channel pruning is analogous to
pruning the heads or activations in the transformer layer. Finally, we could consider pruning foundation models in reinforcement learning for computationally-efficient transfer reinforcement learning \citep{zhu2020transfer} and continual reinforcement learning \citep{caccia2022task}. 

\bibliographystyle{neurips_2022}
\bibliography{ref}

\clearpage
\appendix

\section{Visual support for pruning type}
\label{app:pruning_visual}

\begin{figure}[h]
    \centering
    \begin{subfigure}[b]{0.4\textwidth}
        \includegraphics[width=\textwidth]{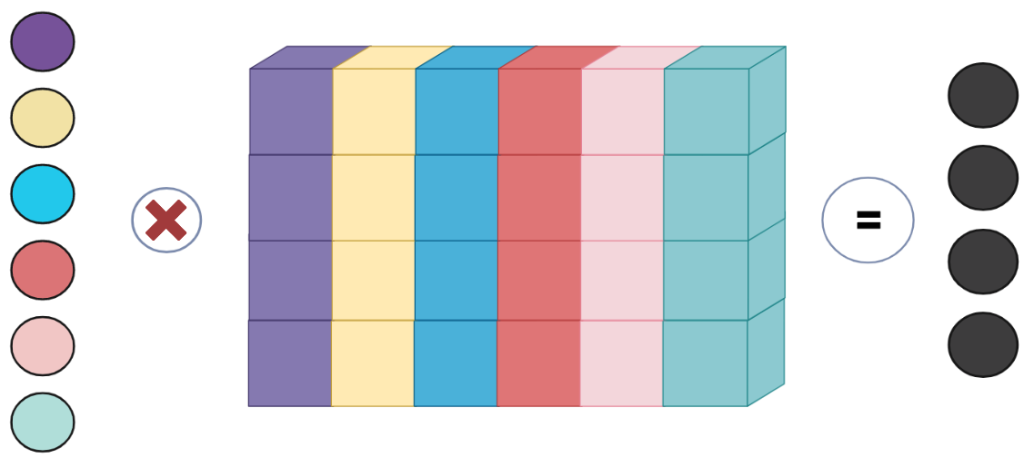}
        \caption{pretrained dense layer}
        \label{fig:viz_ptr}
    \end{subfigure}
    \hspace{0.1\textwidth}
    \begin{subfigure}[b]{0.4\textwidth}
        \includegraphics[width=\textwidth]{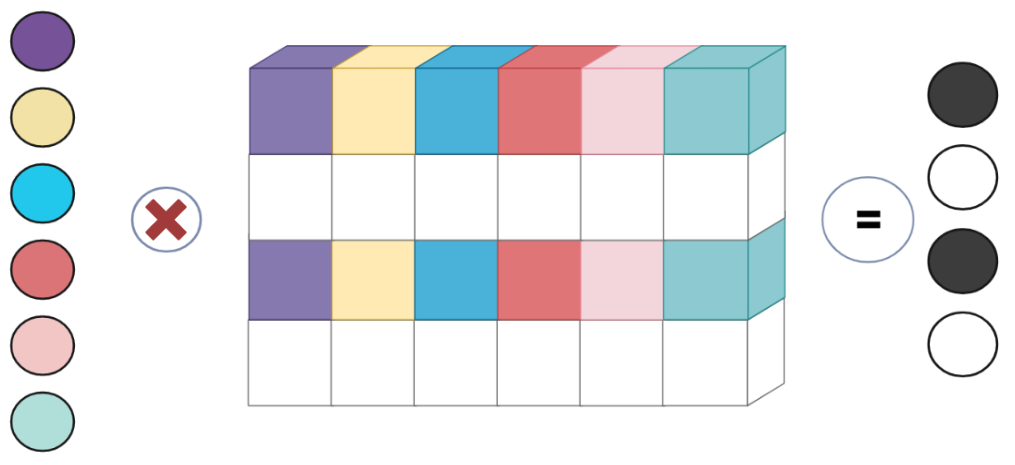}
        \caption{channel pruning}
        \label{fig:viz_channel}
    \end{subfigure}
    \vspace{0.03\textwidth}
    \begin{subfigure}[b]{0.4\textwidth}
        \includegraphics[width=\textwidth]{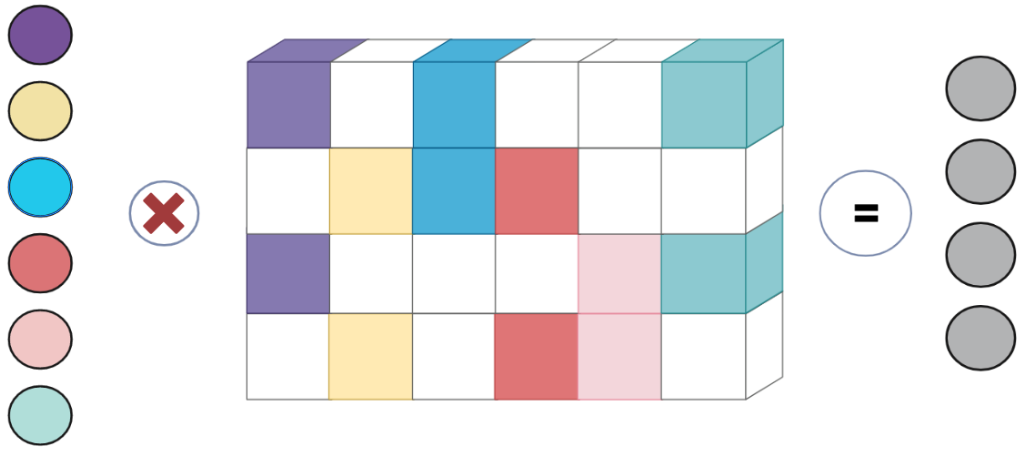}
        \caption{unstructured convolution filter pruning}
        \label{fig:viz_uns_conv}
    \end{subfigure}
    \hspace{0.1\textwidth}
    \begin{subfigure}[b]{0.4\textwidth}
        \includegraphics[width=\textwidth]{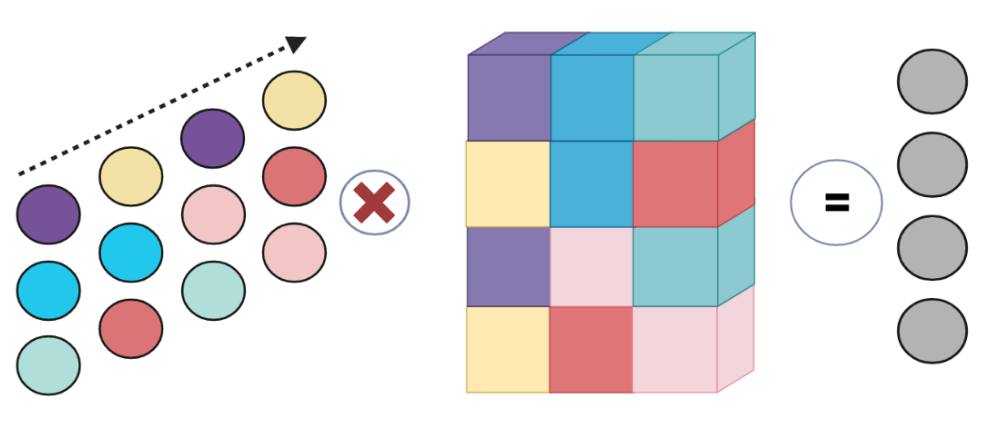}
        \caption{structured convolution filter pruning}
        \label{fig:viz_str_conv}
    \end{subfigure}
    \caption{
        \textbf{Visualizing the Impact of Structured Pruning on Convolution Layers.}
In this illustration, circles represent input/output feature maps ($H \times W$), cubes symbolize convolution kernels ($K \times K$), and the $\otimes$ operator denotes the convolution. The rows and columns of the convolution layer index the output and input channels, respectively. The pruning rate is set at 50\%. The channel pruning approach (b) removes groups of filters responsible for a specific channel, reducing the output representation size and FLOPs by half. In contrast, unstructured convolution filter pruning (c) removes a fraction of filters for each channel while maintaining the output representation size and FLOPs. Structured convolution filter pruning (d) involves restructuring the layer, resulting in a halving of FLOPs, but necessitating the broadcasting of inputs, increasing memory usage. However, the increased memory usage is not prohibitive, as it only occurs locally at the layer level and does not accumulate from one layer to the next.
    } 
    \label{fig:viz_pruning}
\end{figure}

\section{Visual support for parameter efficient methods}
\label{app:paramefficient_visual}

\begin{figure}[h!]
    \centering
    \begin{subfigure}[b]{0.475\textwidth}
        \includegraphics[width=\textwidth]{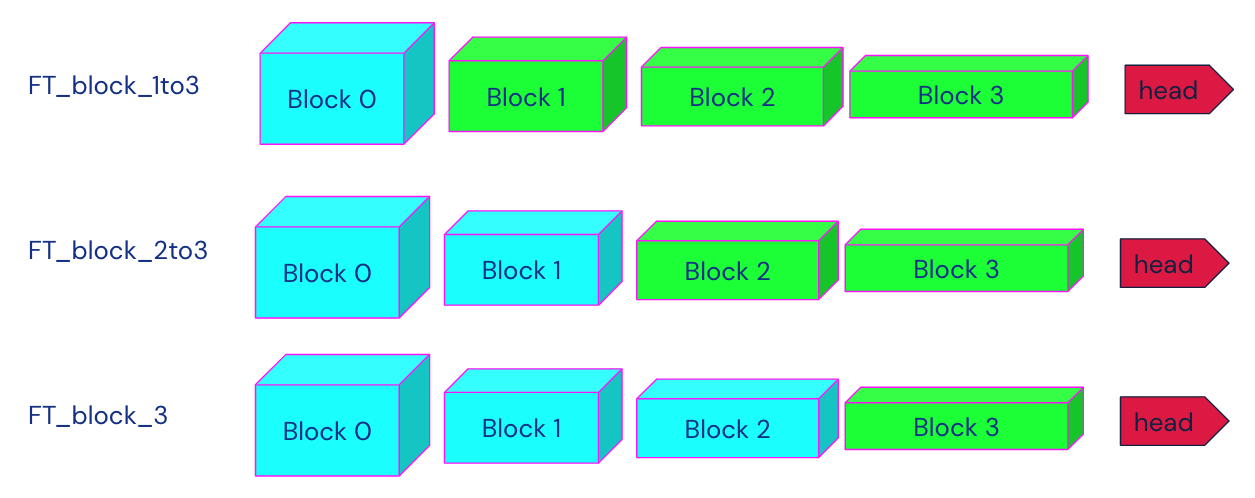}
        \caption{\codeword{ft\_block\_Xto3} baselines}
    \end{subfigure}
    \begin{subfigure}[b]{0.475\textwidth}
        \includegraphics[width=\textwidth]{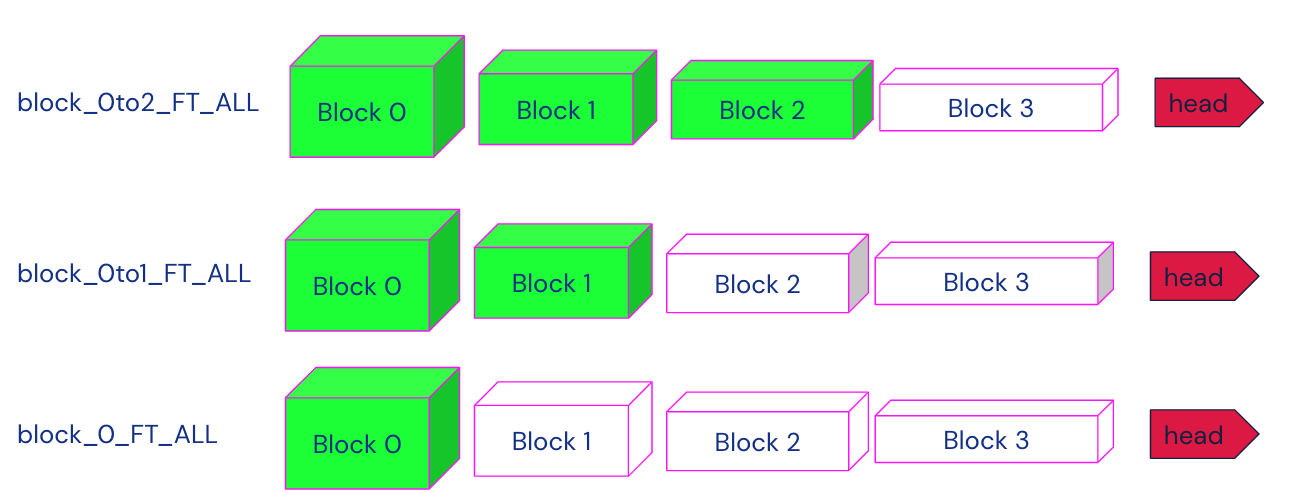}
        \caption{\codeword{block\_0toX\_ft\_all} baselines}
    \end{subfigure}
    \caption{\textbf{Visual support for parameter efficient methods.} The pretrained ResNet is formed of four blocks. Green blocks are trainable, blue are frozen and white are removed.}
\end{figure}

\section{Pruning rate controls the asymptotic performance vs compute efficiency trade-off}
\label{app:sparsity_vals}


\begin{figure}[h!]
    \centering
    \begin{subfigure}[b]{0.475\textwidth}
        \includegraphics[width=\textwidth]{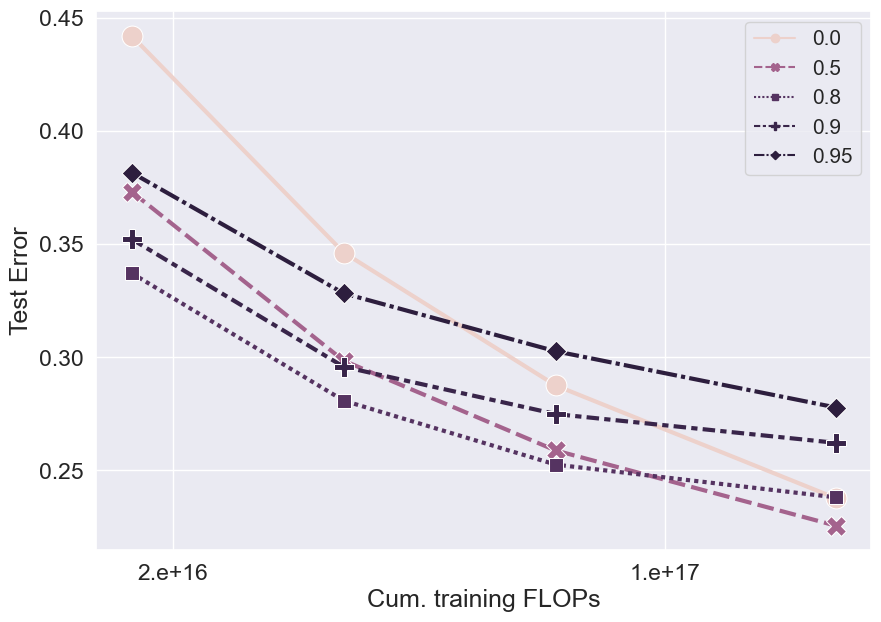}
        \caption{Transfer Learning - Object}
    \end{subfigure}
    \begin{subfigure}[b]{0.475\textwidth}
        \includegraphics[width=\textwidth]{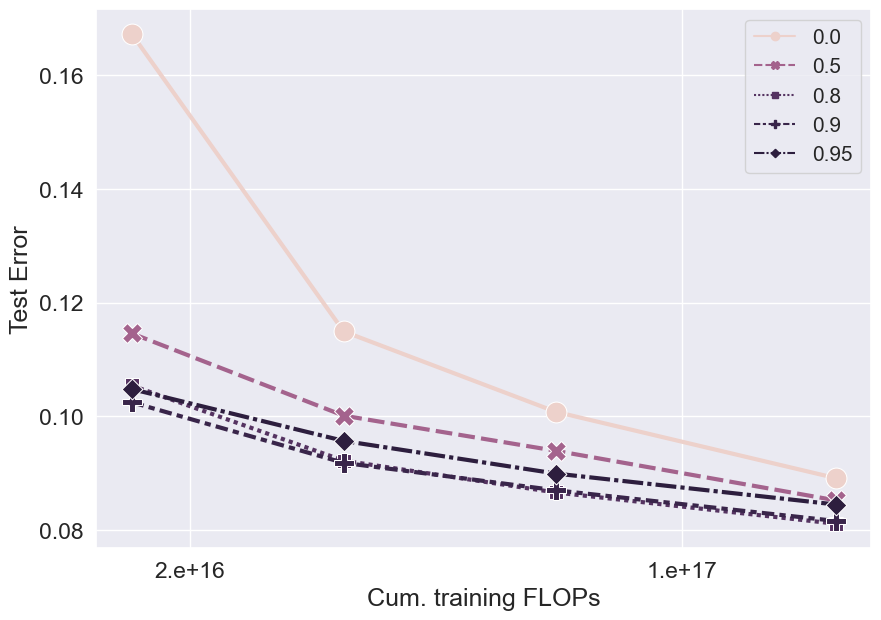}
        \caption{Transfer Learning - Non-object}
    \end{subfigure}
    \caption{\textbf{Pruning rate modulates the asymptotic performance vs computational efficiency trade-off} }
\end{figure}

\section{Pruning signal matters}
\label{app:sparsity_stage}

\begin{figure}[h!]
    \centering
    \begin{subfigure}[b]{0.475\textwidth}
        \includegraphics[width=\textwidth]{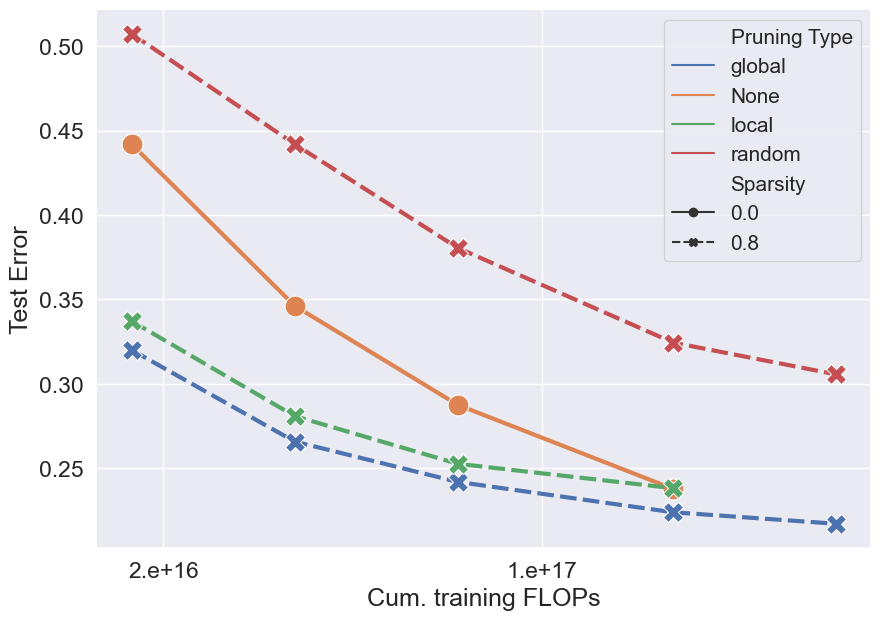}
        \caption{Transfer Learning - Object}
    \end{subfigure}
    \begin{subfigure}[b]{0.475\textwidth}
        \includegraphics[width=\textwidth]{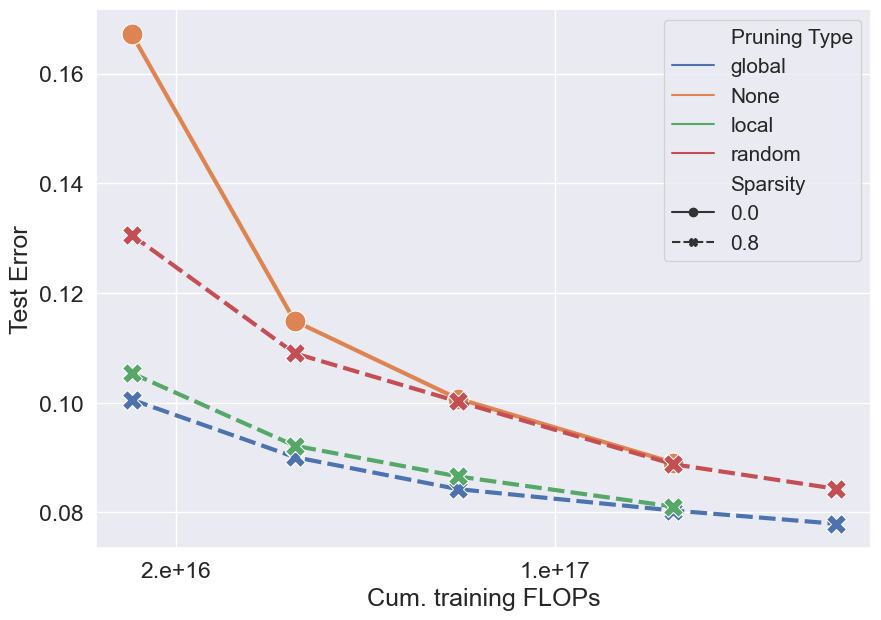}
        \caption{Transfer Learning - Non-object}
    \end{subfigure}
    \caption{\textbf{Pruning signal matters} Randomly pruning the network is not as effective as 
            pruning based on weight magnitude.}
\end{figure}

\end{document}